\documentclass[10pt,twocolumn,letterpaper]{article}

\usepackage{iccv}
\usepackage{times}
\usepackage{epsfig}
\usepackage{graphicx}
\usepackage{amsmath}
\usepackage{amssymb}

\usepackage{booktabs}
\usepackage{enumitem}
\usepackage{makecell}
\usepackage{subcaption}
\usepackage{tabularx}

\usepackage[pagebackref=true,breaklinks=true,letterpaper=true,colorlinks,bookmarks=false]{hyperref}

\iccvfinalcopy 

\def\iccvPaperID{6061} 

\newcommand{\bq}{\mathbf{q}}
\newcommand{\bt}{\mathbf{t}}
\newcommand{\bx}{\mathbf{x}}

\ificcvfinal\fi
\begin{document}

\title{Iterative Transformer Network for 3D Point Cloud}

\author{
Wentao Yuan \qquad David Held \qquad Christoph Mertz \qquad Martial Hebert\\
The Robotics Institute\\
Carnegie Mellon University\\
{\tt\small \{wyuan1, dheld, cmertz, mhebert\}@cs.cmu.edu}
}

\maketitle
\thispagestyle{empty}

\begin{abstract}
    3D point cloud is an efficient and flexible representation of 3D structures. Recently, neural networks operating on point clouds have shown superior performance on 3D understanding tasks such as shape classification and part segmentation. However, performance on such tasks is evaluated on complete shapes aligned in a canonical frame, while real world 3D data are partial and unaligned. A key challenge in learning from partial, unaligned point cloud data is to learn features that are invariant or equivariant with respect to geometric transformations. To address this challenge, we propose the Iterative Transformer Network (IT-Net), a network module that canonicalizes the pose of a partial object with a series of 3D rigid transformations predicted in an iterative fashion. We demonstrate the efficacy of IT-Net as an anytime pose estimator from partial point clouds without using complete object models. Further, we show that IT-Net achieves superior performance over alternative 3D transformer networks on various tasks, such as partial shape classification and object part segmentation. Our code and data are available at \url{https://github.com/wentaoyuan/it-net}.
\end{abstract}

\begin{figure}[!t]
    \centering
    \includegraphics[width=0.9\linewidth]{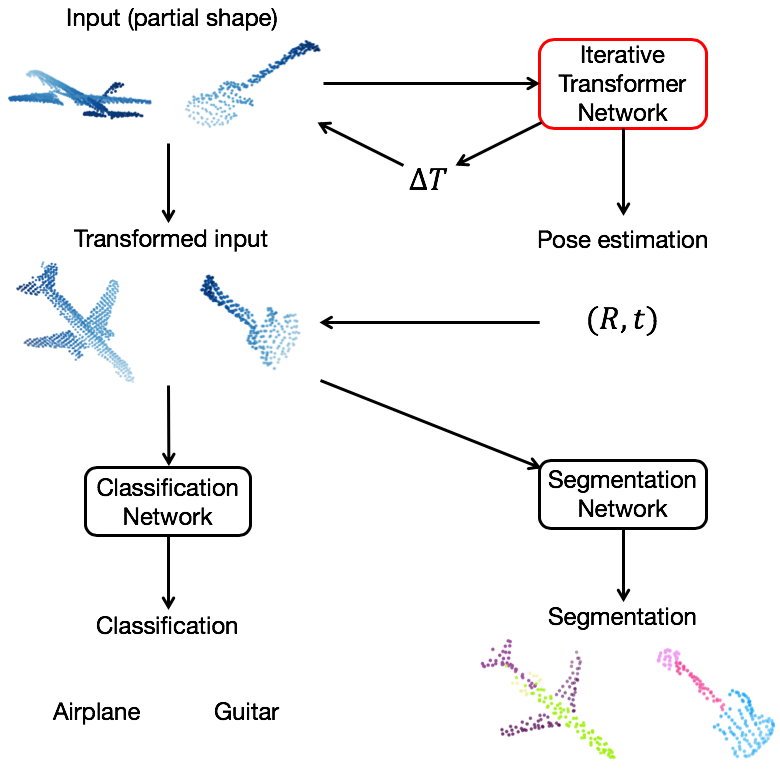}
    \caption{Iterative Transformer Network (IT-Net) predicts rigid transformations from partial point clouds in an iterative fashion. It can be used independently as a pose estimator or jointly with classification and segmentation networks.}
    \label{fig:teaser}
\end{figure}

\section{Introduction} \label{sec:intro}
3D point cloud is the raw output of most 3D sensors and multiview stereo pipelines \cite{furukawa2010accurate} and a widely used representation for 3D structures in applications such as autonomous driving \cite{geiger2012we} and augmented reality \cite{klein2007parallel}. Due to its efficiency and flexibility, there is a growing interest in using point clouds for high level tasks such as object recognition, skipping the need for meshing or other post-processing. These tasks require an understanding of the semantic concept represented by the points. On other modalities like images, deep neural networks \cite{he2016deep,krizhevsky2012imagenet} have proven to be a powerful model for extracting semantic information from raw sensor data, and have gradually replaced hand-crafted features. A similar trend is happening on point clouds. With the introduction of deep learning architectures like PointNet \cite{qi2017pointnet}, it is possible to train powerful feature extractors that outperform traditional geometric descriptors on tasks such as shape classification and object part segmentation.

\begin{figure}[!t]
    \centering
    \includegraphics[width=\linewidth]{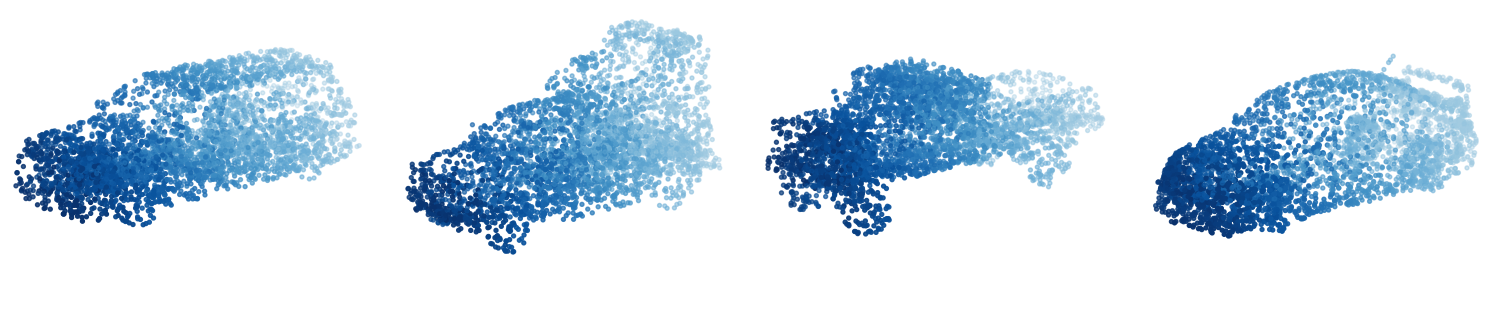}
    \includegraphics[width=\linewidth]{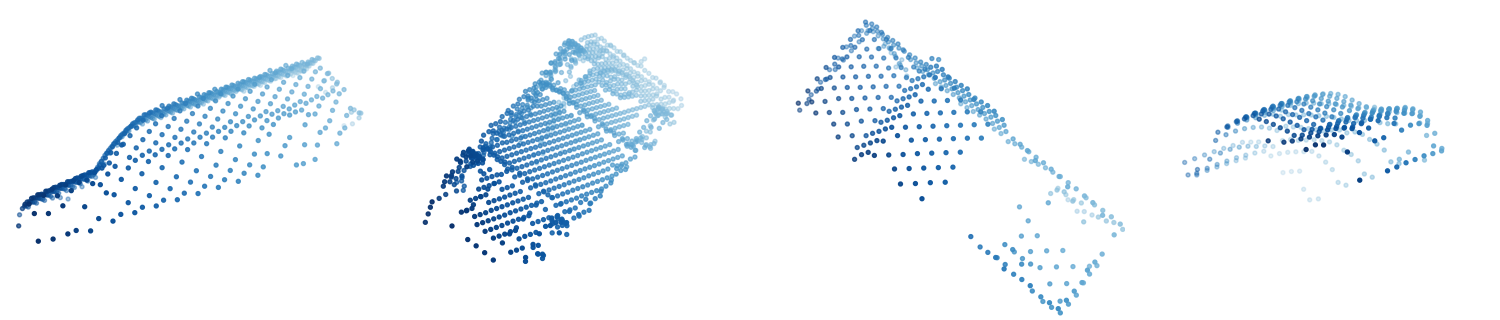}
    \caption{Complete point clouds aligned in a canonical frame from ShapeNet (top row) versus partial, unaligned point clouds from our dataset (bottom row).} 
    \label{fig:canonical}
\end{figure}

However, existing benchmark datasets \cite{wu20153d,yi2016scalable} that are used to evaluate performance on these tasks make two simplifying assumptions: first, the point clouds are sampled from complete shapes; second, the shapes are aligned in a canonical coordinate system\footnote{In ModelNet \cite{wu20153d}, shapes are allowed to have rotations, but only along the vertical axis.} (see Figure \ref{fig:canonical}). These assumptions are rarely met in real world scenarios. First, due to occlusions and sensor limitations, real world 3D scans usually contain missing regions. Second, point clouds are often obtained in the sensor’s coordinates, which do not align with the canonical coordinates of the object model. In other words, real 3D point cloud data are \emph{partial} and \emph{unaligned}.

In this work, we tackle the problem of learning from partial, unaligned point cloud data. To this end, we build a dataset consisting of partial point clouds generated from virtual scans of CAD models in ModelNet \cite{wu20153d} and ShapeNet \cite{chang2015shapenet} as well as real world scans from ScanNet \cite{dai2017scannet}. Our dataset contains challenging inputs with arbitrary 3D rotation, translation and realistic self-occlusion patterns.

A key challenge in learning from such data is how to learn features that are invariant or equivariant with respect to geometric transformations. For tasks like classification, we want the output to remain the same if the input is transformed. This is called invariance. For tasks like pose estimation, we want the output to vary according to the transformation applied on the input. This is called equivariance. One way to achieve invariance or equivariance is via a transformer network \cite{jaderberg2015spatial}, which predicts a transformation that is applied to the input before feature extraction. The predicted transformation allows explicit geometric manipulation of data within the network so the inputs can be aligned into a canonical space that makes subsequent tasks easier.

\begin{figure}[!t]
    \centering
    \begin{subfigure}{0.49\linewidth}
        \includegraphics[width=\linewidth]{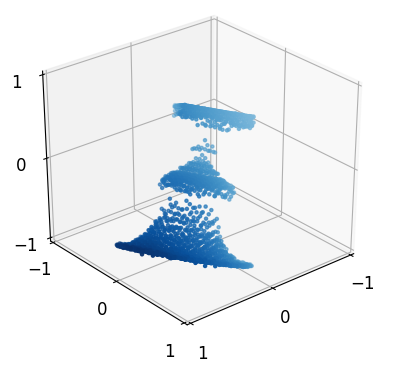}
        \caption{Input}
    \end{subfigure}
    \begin{subfigure}{0.49\linewidth}
        \includegraphics[width=\linewidth]{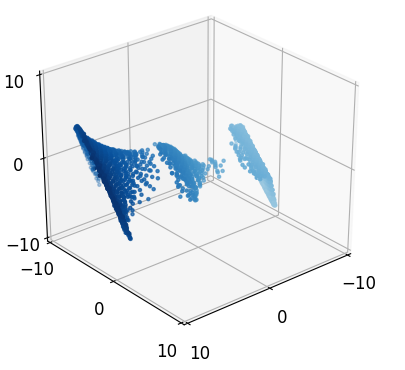}
        \caption{Transformed input}
    \end{subfigure}
    \caption{T-Net \cite{qi2017pointnet} scales and distorts the input shape (a vase). Note the different scales on the plots.}
    \label{fig:distortion}
\end{figure}

T-Net \cite{qi2017pointnet} is a transformer network based on PointNet that operates on 3D point clouds. However, T-Net outputs an unconstrained affine transformation. This can introduce undesirable shearing and scaling which causes the object to lose its shape (see Figure \ref{fig:distortion}). Moreover, T-Net is evaluated on inputs with 2D rotations only.

To address the shortcomings of T-Net, we propose a novel transformer network on 3D point clouds, named Iterative Transformer Network (IT-Net). IT-Net has two major differences from T-Net. First, it outputs a rigid transformation instead of an affine transformation. Outputting rigid transformation allows the outputs to be used directly as estimates for object poses and leads to better performance on subsequent tasks such as shape classification and part segmentation. Second, instead of predicting the transformation in a single step, IT-Net takes advantage of an iterative refinement scheme which decomposes a large transformation into smaller ones that are easier to predict. The multi-step output not only increases accuracy of the predicted transformation, but also allows anytime prediction, i.e. the result can be gradually refined until the test-time computational budget is depleted. 

We demonstrate the advantage of IT-Net over alternative transformer networks on three point cloud learning tasks -- pose estimation, shape classification and part segmentation (Figure \ref{fig:teaser}) -- with partial, unaligned inputs from synthetic as well as real world 3D data. 

The key contributions of our work are as follows:
\begin{itemize}[itemsep=0pt]
    \item We propose a novel transformer network called IT-Net that adds geometric invariance/equivariance to networks operating on 3D point clouds;
    \item We demonstrate that IT-Net can be used as an anytime pose estimator which outperforms strong baselines when applied to point cloud alignment;
    \item We show that IT-Net outperforms existing transformer networks on point clouds when trained jointly with various classification or segmentation networks;
    \item We introduce a new dataset for pose estimation, shape classification and part segmentation consisting of partial, unaligned point clouds.
\end{itemize}

\section{Related Work}
\paragraph{Feature Learning on Point Clouds}
Traditional point feature descriptors \cite{rusu2009fast,sun2009concise} rely on geometric properties of points such as curvatures. They do not encode semantic information and it is non-trivial to find the combination of features that is optimal for specific tasks.

Qi \etal \cite{qi2017pointnet} introduces a way to extract semantic and task-specific features from point clouds using a neural network, which outperforms competing methods on several shape analysis tasks like shape classification. Subsequent works \cite{li2018pointcnn,qi2017pointnet++,wang2018dynamic} further improves the performance of point cloud-based networks by accounting for interactions among local neighborhoods of points.


\paragraph{Spatial Transformer Network}
Spatial Transformer Network (STN) \cite{jaderberg2015spatial} is a network module that performs explicit geometric transformations on the input image in a differentiable way.
STN introduces invariance to geometric transformations and can be trained jointly with various task-specific networks to improve their performance.

IC-STN \cite{lin2016inverse} is an extension of STN that makes use of an iterative scheme inspired by the Lucas-Kanade algorithm \cite{lucas1981iterative}. Our network utilizes a similar iterative scheme to predict accurate geometric transformations.

\begin{figure*}
    \centering
    \includegraphics[width=\linewidth]{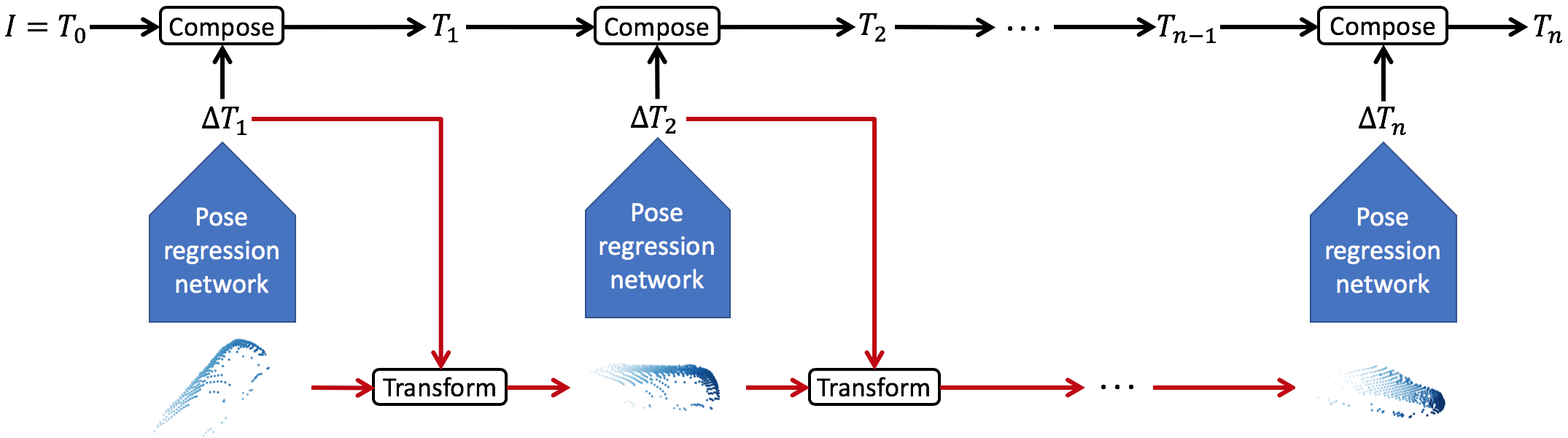}
    \caption{Illustration of the iterative scheme employed by IT-Net. At each iteration, the output of the pose regression network is used to transform the input for the next iteration. The parameters of the pose regression network shown in blue arrows are shared across iterations. The final output is a composition of the transformations predicted at each iteration. Arrows colored in red indicate places where the gradient flow is stopped to decorrelate the inputs at different iterations (see Sec. \ref{sec:detail}).}
    \label{fig:arch}
\end{figure*}

\paragraph{Iterative Error Feedback}
The idea of using iterative error feedback (IEF) in neural networks have been studied in the context of 2D human pose estimation \cite{carreira2016human} and taxonomic prediction \cite{zamir2017feedback}. Under the IEF framework, instead of trying to directly predict the target in a feed-forward fashion, the network predicts the error in the current estimate and corrects it iteratively. While our proposed network falls under this general framework, unlike previous works, it does not use intermediate supervision or separate stages of training. Rather, the loss is applied at a certain iteration during training and the gradient is propagated through the composition of outputs from previous iterations.

\section{Iterative Transformer Network} \label{sec:arch}
Iterative Transformer Network (IT-Net) takes a 3D point cloud and produces a transformation that can be used directly as a pose estimate or applied to the input before feature extraction for subsequent tasks. 
IT-Net has two key features that differentiate it from existing transformer networks on 3D point clouds: 1) it predicts a 3D rigid transformation; 2) the final output is composed of multiple transformations produced in an iterative fashion.

\subsection{Rigid Transformation Prediction} \label{sec:rigid}
The output of IT-Net is a 3D rigid transformation $T$, consisting of a rotation $R$ and translation $\bt$ where $R$ is a $3\times3$ matrix satisfying $RR^T=I,\det(R)=1$ and $\bt$ is a $3\times1$ vector. Due to the constraints on $R$, it is inconvenient to represent the rotation as a $3\times3$ matrix during optimization. Thus, many classical \cite{besl1992method} as well as modern deep learning methods \cite{kendall2015posenet,xiang2017posecnn} parametrize 3D rotations with unit quaternions. The quaternion parametrization allows us to map an arbitrary 4D vector to a valid 3D rotation.

A single iteration of IT-Net is a pose regression network that takes a point cloud and outputs 7 numbers parametrizing a 3D rigid transformation. The first 4 numbers are normalized into a unit quaternion $\bq$ and the last 3 are treated as a 3D translation vector $\bt$. Then, $\bq$ and $\bt$ are assembled into a $4\times4$ matrix $T=\begin{bmatrix}R(\bq)&\bt\\\mathbf{0}&1\end{bmatrix}$ where $R(\bq)$ is the rotation matrix corresponding to $\bq$. The matrix representation turns the composition of two rigid transformations into a matrix multiplication, which is convenient for composing the outputs from multiple iterations. We use PointNet \cite{qi2017pointnet} as the regression network for its simplicity, but other point cloud-based networks can be used as well.

In contrast to the affine transformation produced by T-Net \cite{qi2017pointnet}, the rigid transformation predicted by IT-Net can be directly interpreted as a 6D pose, making it possible to use IT-Net independently for pose estimation. More importantly, rigid transformations preserve scales and angles. As a result, the appearance of a point cloud will not vary drastically if it is transformed by the output of IT-Net. This makes it possible to apply the same network iteratively to obtain a more accurate estimation of the transformation.

We note that it is possible to add a regularization term $\|AA^T-I\|$ that forces an affine matrix $A$ to be orthogonal in order to achieve similar effects of predicting a rigid transformation\footnote{In \cite{qi2017pointnet}, this regularization is added to the feature transformation, but not to the input transformation.}. However, the constraint, no matter how close to satisfied, cannot produce a truly rigid transformation that prevents the deformation of inputs. As shown in Sec. \ref{sec:cls}, the results of the regularized network are not as good as the network that directly outputs rigid transformations.

\subsection{Iterative Alignment} \label{sec:iter}
The idea of using an iterative scheme for predicting geometric transformations goes back to the classical Lucas-Kanade (LK) algorithm \cite{lucas1981iterative} for estimating dense alignment between images. The key insight of LK is that the complex non-linear mapping from image appearance to geometric transformations can be estimated iteratively using simple linear predictors. Specifically, at each iteration, a warp transformation $\Delta p$ is predicted with a linear function that takes a source and a target image as inputs. Then, the source image is warped by $\Delta p$ and the process is repeated. The final transformation is a composition of $\Delta p$ at each step. Later, \cite{baker2004lucas} shows that the parameters used to predict $\Delta p$ can remain constant across iterations while achieving the same effect as non-constant predictors.

The same idea is employed in the Iterative Closest Point (ICP) algorithm \cite{besl1992method} for the alignment of 3D point clouds. At each iteration of ICP, a corresponding set is identified and a rigid transformation $\Delta T$ is produced to align the corresponding points. Then, the source point cloud is transformed by $\Delta T$ and the process is repeated. Again, the final output is a composition of $\Delta T$ at each step. The effectiveness of ICP shows that the iterative refinement framework applies not only to images, but also to 3D point clouds.

The multi-iteration IT-Net (Figure \ref{fig:arch}) can be viewed as an instantiation of this iterative framework. Specifically, the prediction of the transformation $T$ is unfolded into multiple iterations. At the $i$-th iteration, a rigid transformation $\Delta T_i$ is predicted as described in Sec. \ref{sec:rigid}. Then, the input is transformed by $\Delta T_i$ and the process is repeated. The final output after $n$ iterations is a composition of the transformations predicted at each iteration, which can be written as a simple matrix product $T_n=\prod_{i=1}^n\Delta T_i$.

We use a fixed predictor (i.e. share the network's parameters) across iterations following \cite{baker2004lucas}. In addition to reduction in the number of parameters, the fixed predictor allows us to use different numbers of unfolded iterations in training and testing. As will be shown in Sec. \ref{sec:pose}, once trained, IT-Net can be used as an anytime predictor where increasingly accurate pose estimates can be obtained as the network is applied for more iterations.

The iterative scheme can be interpreted as a way to automatically generate a curriculum, which breaks down the original task into a set of simpler pieces. In earlier iterations, the network learns to predict large transformations that bring the input near its canonical pose. In later iterations, the network learns to predict small transformations that adjusts the estimate from previous iterations. Note that the curriculum is not manually defined but rather generated by the network itself to optimize the end goal. It will be empirically shown in Sec. \ref{sec:pose} that this curriculum emerges from the training of IT-Net.



\subsection{Implementation Details} \label{sec:detail}
In addition to the key ingredients above, there are a couple of details that are important for the training of IT-Net. First, we initialize the network to predict the identity transformation $\bq=[1\;0\;0\;0]$, $\bt=[0~0~0]$. In this way, the default behavior of each iteration is to preserve the transformation predicted by previous iterations. This identity initialization helps prevent the network from producing large transformations which cancel each other. Second, we stop the gradients propagating through input transformations (red arrows in Figure \ref{fig:arch}) and let the gradients propagate through the output composition only (black arrows in Figure \ref{fig:arch}). This removes dependency among inputs at different iterations which leads to gradient explosion. Empirical evaluations for these design choices can be found in Sec. \ref{sec:pose}.

\section{Experiments}
We evaluate IT-Net on various point cloud learning tasks. In Sec. \ref{sec:pose}, we demonstrate the ability of IT-Net to estimate the canonical pose of an object from partial views in an anytime fashion. In Sec. \ref{sec:cls}, we show that IT-Net outperforms existing transformer networks when trained jointly with state-of-the-art classifiers on partial, unaligned shapes from both synthetic and real world data. In Sec. \ref{sec:seg}, we test IT-Net's capability to improve performance of state-of-the-art models on object part segmentation, showing that the invariance learned by IT-Net can benefit a variety of shape analysis tasks. We implemented all our networks in TensorFlow \cite{abadi2016tensorflow}. Detailed hyperparameter settings can be found in the supplement.

\paragraph{Dataset}
To evaluate the performance of point cloud learning tasks under a more realistic setting, we build a dataset of object point clouds which captures the incomplete and unaligned nature of real world 3D data. The dataset consists of the following parts:
\begin{itemize}
    \item \textbf{Partial ModelNet40} includes 81,212 object point clouds in 40 categories generated from ModelNet40 \cite{wu20153d}, split into 78,744 for training and 2,468 for testing. Each point cloud is generated by fusing up to 4 depth scans of a CAD model into a point cloud. 
    \item \textbf{ScanNet Objects} consists of 9,122 object point clouds in 33 categories collected from ScanNet \cite{dai2017scannet}, split into 8,098 for training and 1,024 for testing. The point clouds are obtained by cropping indoor RGBD scans with labeled bounding boxes, where realistic sensor noise and clutter in the RGBD scans are kept.
    \item \textbf{ShapeNet Pose} includes 24,000 object point clouds in the car and chair category, generated by back-projecting depth scans of 4,800 ShapeNet \cite{chang2015shapenet} models from uniformly sampled viewpoints into the camera's coordinates. Each point cloud is labeled with the transformation that aligns it to the model's coordinates. The data are split into training, validation and testing with a 10:1:1 ratio. Note that the test set and the training set are created with different object models.
    \item \textbf{ShapeNet Part} contains 16,881 object point clouds in 16 categories from ShapeNet \cite{chang2015shapenet}, split into 13,937 for training and 2,874 for testing. Each point cloud is labeled with 2-6 parts using the labels provided in \cite{yi2016scalable}. Since the part labels are provided for point clouds and not meshes, we use an approximate rendering procedure that mimics an orthographic depth camera to create realistic-looking partial point clouds.
\end{itemize}
More details on data generation are in the supplement.

\subsection{Object Pose Estimation} \label{sec:pose}
We investigate the efficacy of the iterative refinement scheme on the task of estimating the canonical pose of an object from a partial observation. Specifically, we use IT-Net to predict the transformation that aligns the input shape to a canonical frame defined across all models in the same category (see the top row in Figure \ref{fig:canonical}). Unlike most existing works on pose estimation, we do not assume knowledge of the complete object model and we train a single network that generalizes to different objects in the same category.

The network architecture is described in Sec. \ref{sec:arch} where the pose regression network is a PointNet \cite{qi2017pointnet}. Details about the number of layers and parameters in the pose regression network can be found in the supplement.

An explicit loss is applied to the output transformation. For the loss function, we use a variant of PLoss proposed in \cite{xiang2017posecnn}, which measures the average distance between the same set of points under the estimated pose and the ground truth pose. Compared to the L2 loss used in earlier works \cite{kendall2015posenet}, this loss has the advantage of automatically handling the tradeoff between small rotations and small translations. The loss can be written as
\begin{equation}
    L((R,\bt),(\widetilde{R},\tilde{\bt})) = \frac{1}{|X|}\sum_{\bx\in X}\|(R\bx+\bt)-(\widetilde{R}\bx+\tilde{\bt})\|_2^2,
\end{equation}
where $R,\bt$ are the ground truth pose and $\widetilde{R},\tilde{\bt}$ are the estimated pose and $X$ is the set of input points.

We trained IT-Net under different settings and evaluated their performance on the car point clouds in ShapeNet Pose. Similar experiments on the chair point clouds in ShapeNet Pose can be found in the supplement. In what follows, we provide detailed analysis of the results.

\paragraph{Number of unfolded iterations}
The number of unfolded iterations during training can be treated as a hyperparameter that controls the iteration at which the loss is applied. We trained IT-Net with different number of unfolded iterations and, as shown in Table \ref{tab:pose_abl}, IT-Net trained with 5 unfolded iterations gives the best performance.

\begin{figure*}
    \centering
    \begin{subfigure}{\linewidth}
        \centering
        \begin{tabular}{ccc}
            Iteration 0 & Iteration 3 & Iteration 6 \\
            \includegraphics[width=0.3\linewidth]{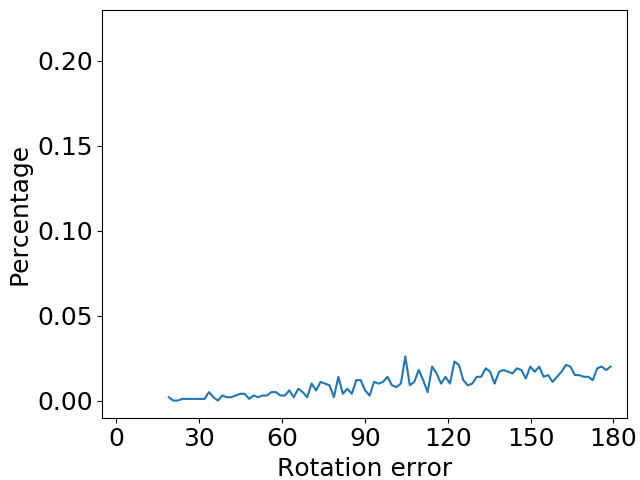} & \includegraphics[width=0.3\linewidth]{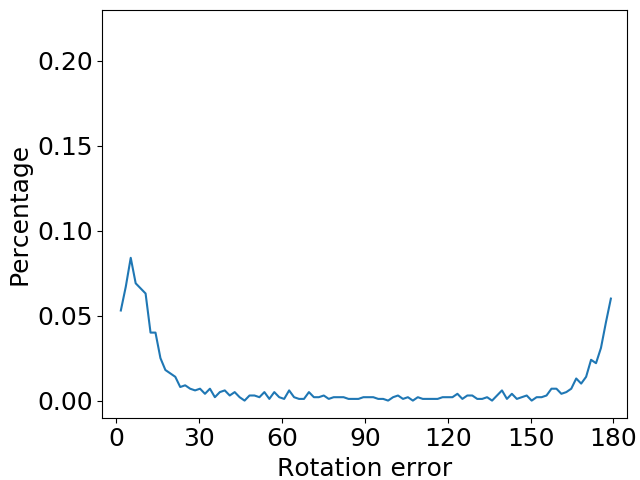} & \includegraphics[width=0.3\linewidth]{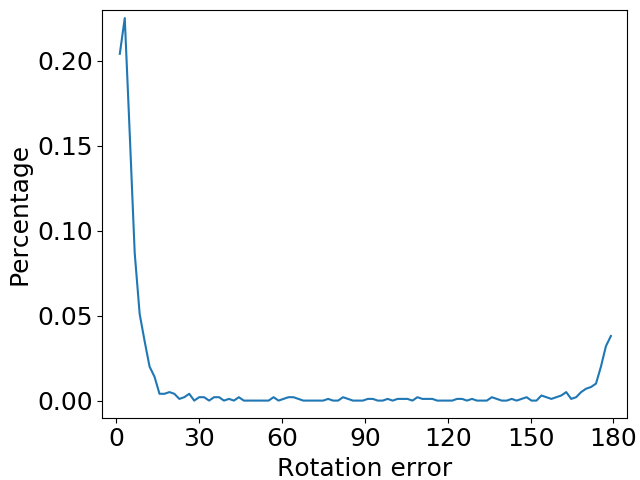} \\
            \includegraphics[width=0.3\linewidth]{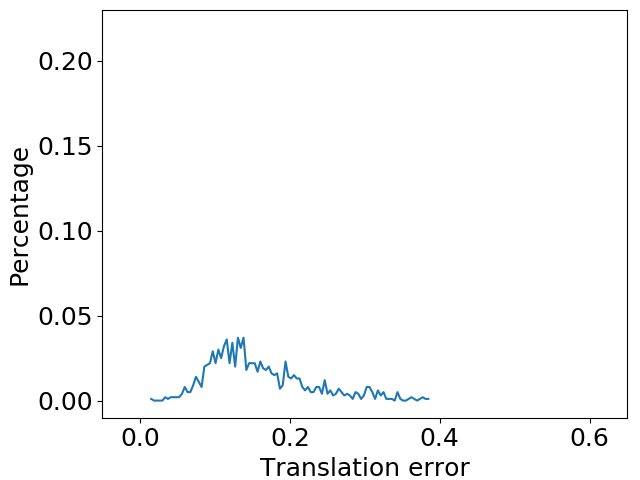} & \includegraphics[width=0.3\linewidth]{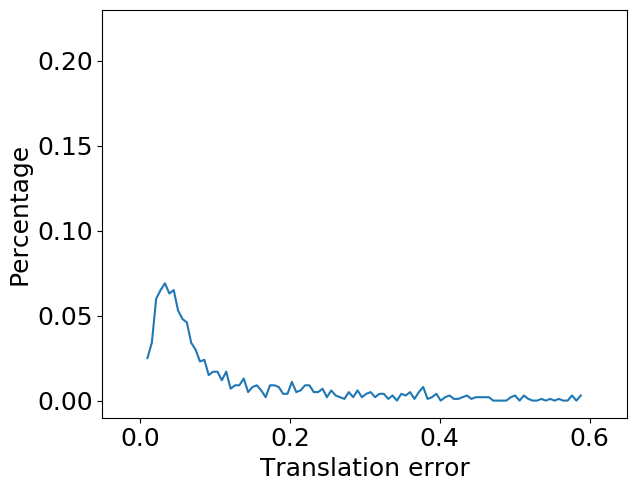} & \includegraphics[width=0.3\linewidth]{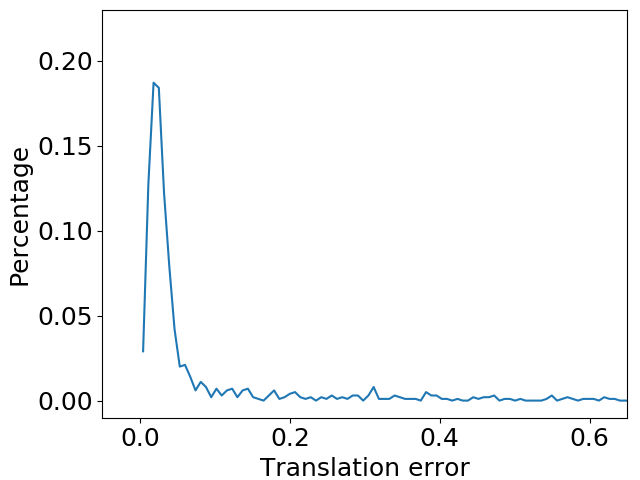} \\
        \end{tabular}
        \caption{Distribution of rotation and translation errors at different iterations}
        \label{fig:pose_agg}
    \end{subfigure} \\
    \vspace{0.4cm}
    \begin{subfigure}{\linewidth}
        \begin{tabular}{p{1.7cm}p{1.7cm}p{1.7cm}p{2cm}|p{1.7cm}p{1.7cm}p{1.7cm}p{2cm}}
            \centering Input & Iteration 3 & Iteration 6 & Ground truth & \centering Input & Iteration 3 & Iteration 6 & Ground truth \\
            \multicolumn{4}{c|}{\includegraphics[width=8cm]{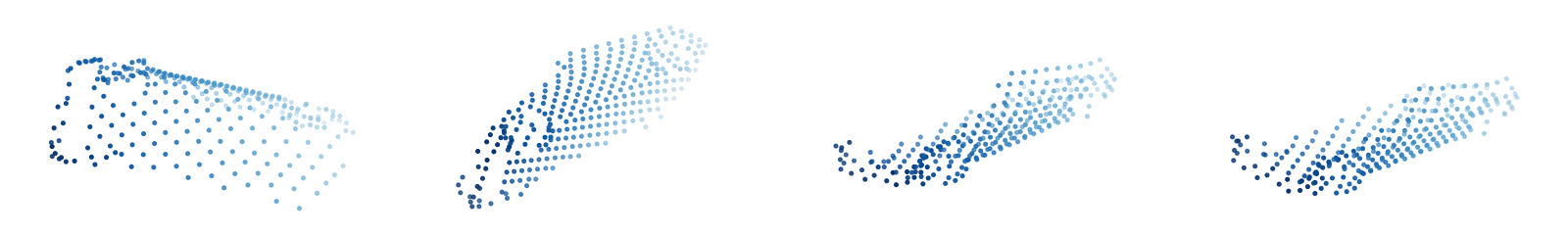}} & \multicolumn{4}{c}{\includegraphics[width=8cm]{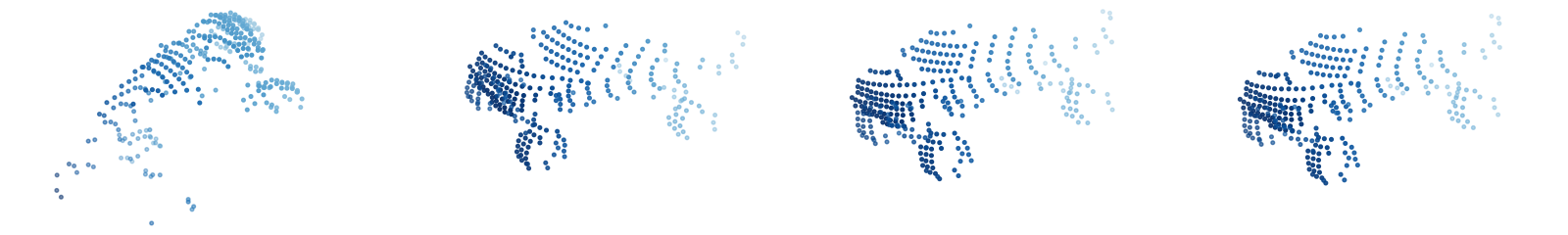}} \\
            \multicolumn{4}{c|}{\includegraphics[width=8cm]{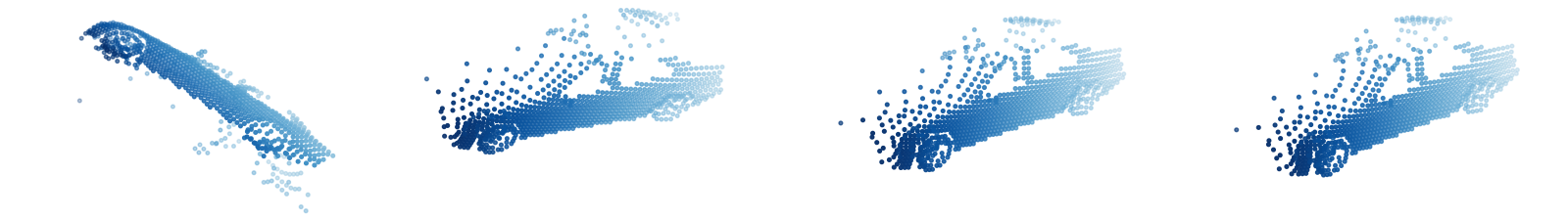}} & \multicolumn{4}{c}{\includegraphics[width=8cm]{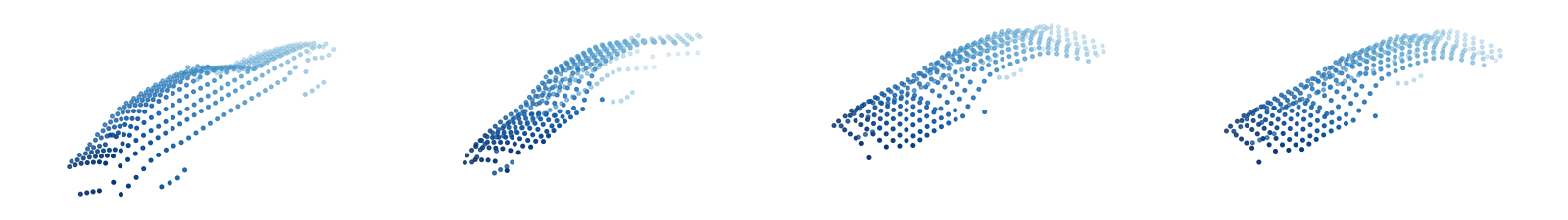}} \\
        \end{tabular}
        \caption{Qualitative examples}
    \end{subfigure}
    \caption{\textbf{(a)} The distribution (PDF) of rotation and translation error of 2,400 test instances at different iterations. Note how the error distribution skews towards 0 in later iterations. The peak at 180 degrees for rotation error is caused by symmetries in the car models. \textbf{(b)} Qualitative results at corresponding iterations.}
    \label{fig:pose}
\end{figure*}

In Figure \ref{fig:pose_agg}, we visualized the distribution of input poses at different iterations during training by measuring errors with respect to the canonical pose. It can be observed that the distribution of poses skews towards the canonical pose in later iterations. This is evidence that the network generates a curriculum as mentioned in Sec. \ref{sec:iter}. We observe that an appropriate distribution of examples in the generated curriculum is key to good performance. With too few unfolded iterations, the network does not see enough examples with small errors and thus fails to predict accurate refinements when the shape is near its canonical pose. With too many unfolded iterations, the network sees too many examples with small errors, overfits to them and becomes too conservative in its prediction. Empirically, 5 iterations turns out to be a good compromise.

\begin{table}
    \centering
    \small
    \begin{tabular}{cccccccc}
        \toprule
        \thead{Unfolded \\ iterations} & 1 & 2 & 3 & 4 & 5 & 6 & 7 \\
        \midrule
        no init & 17.1 & 0.5 & 18.4 & 7.3 & 6.2 & 41.1 & 13.1 \\
        no stop & 36.0 & 5.0 & 0.0 & 0.0 & 4.1 & 4.1 & 0.0 \\
        ours & 36.9 & 41.7 & 47.7 & 61.1 & 67.6 & 62.6 & 48.2 \\
        \bottomrule
    \end{tabular}
    \caption{Pose accuracy (\%) with error threshold $10^\circ,0.1$ of IT-Nets with different number of unfolded iterations during training. No init means the output is \emph{not} initialized as the identity transformation. No stop means the gradient is \emph{not} stopped during input transformations.}
    \label{tab:pose_abl}
\end{table}

\begin{figure}
    \centering
    \includegraphics[width=0.8\linewidth]{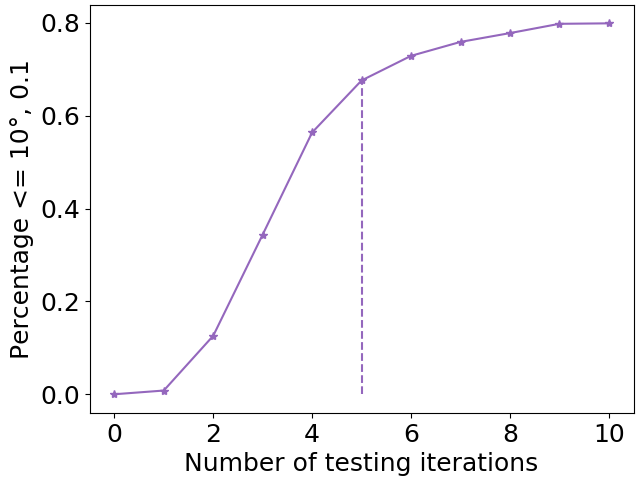}
    \caption{Pose accuracy (\%) against the number of iterations applied during inference. The dotted line corresponds to the number of unfolded iterations in training. Note how the accuracy keeps improving even when more iterations are applied than the network is trained for.}
    \label{fig:pose_any}
\end{figure}

\begin{figure*}
    \centering
    \includegraphics[width=0.32\linewidth]{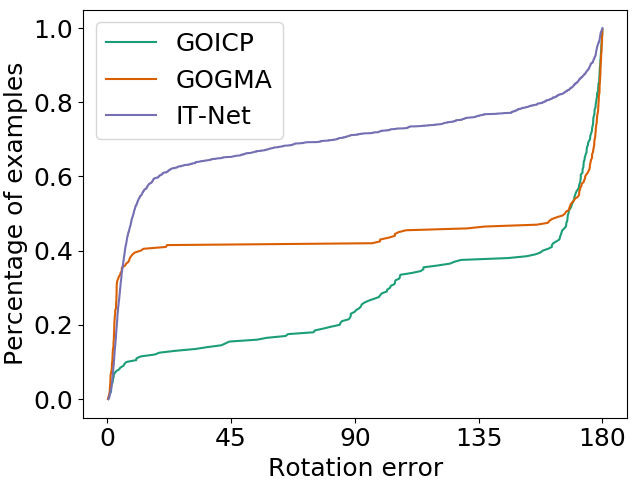}
    \includegraphics[width=0.32\linewidth]{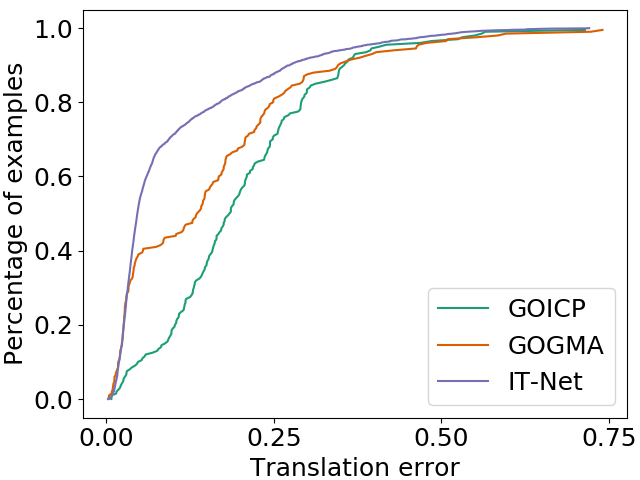}
    \includegraphics[width=0.32\linewidth]{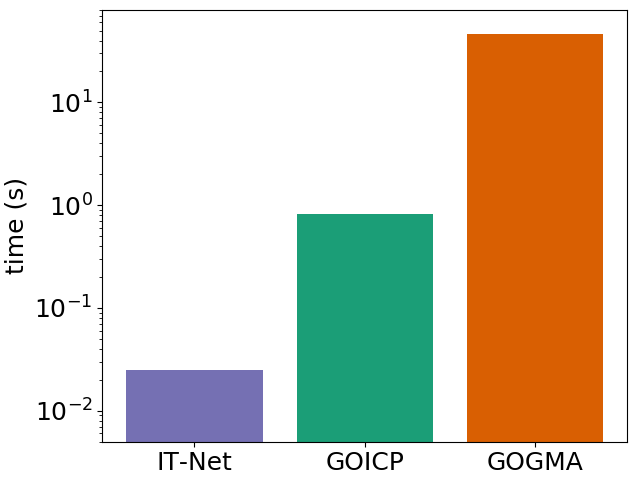}
    \caption{Comparison with non-learning baselines on point cloud alignment. The two plots on the left show the CDF of rotation and translation errors over 1000 test instances. The plot on the right shows the average running time per instance.}
    \label{fig:regis}
\end{figure*}

\paragraph{Anytime pose estimation}
As noted in Sec. \ref{sec:iter}, sharing weights across iterations allows us to use a different number of iterations during inference than during training. Figure \ref{fig:pose_any} shows the pose accuracy of an IT-Net trained with 5 unfolded iterations against the number of iterations applied during inference. It can be seen that the performance keeps increasing as the number of iteration increases. This property allows us to use IT-Net as an anytime pose estimator. In other words, during inference, we can keep applying the trained IT-Net to obtain increasingly accurate pose estimates until the time budget runs out (each iteration takes about 0.025s on a 3.60GHz Intel Core i7 CPU).

\paragraph{Comparison with non-learning baselines}
We applied our pose estimation network to the problem of point cloud alignment and compared the results with classical baselines that is not learning-based. Specifically, for each pair of shapes in the test set, we computed their relative transformation using the poses predicted by IT-Net. The results are compared against two state-of-the-art, non-learning-based alignment methods, GOICP \cite{yang2013go} and GOGMA \cite{campbell2016gogma}. Unlike classical ICP which only works with good initialization, these baseline methods can estimate alignment from arbitrary initialization. The results are shown in Figure \ref{fig:regis}. Note that IT-Net not only produces more accurate alignment, but is also several orders of magnitude faster on average. The running times are measured on a 3.60GHz Intel Core i7 CPU.

\paragraph{Ablation studies}
We conducted ablation studies to validate our design choices described in Sec. \ref{sec:detail}, i.e. initializing the network's prediction with the identity transformation and stopping the gradient flow through input transformations. The results are summarized in Table \ref{tab:pose_abl}. It can be seen that the performance degrades significantly without either identity initialization or gradient stopping, which indicates that both are crucial for the iterative refinement scheme to achieve desired behavior.

\subsection{3D Shape Classification} \label{sec:cls}
The network used for the partial shape classification task consists of two parts -- the transformer and the classifier. The transformer takes a point cloud and produces a transformation $T$. The classifier takes the point cloud transformed by $T$ and outputs a score for each class. The entire network is trained with cross-entropy loss on the class scores and no explicit supervision is applied on the transformation $T$.

We compare classifiers trained with three different transformers, IT-Net, T-Net and regularized T-Net (T-Net reg). The transformers share the same architecture except for the last output layer. IT-Net outputs 7 numbers for rotation (quaternion) and translation; T-Net and T-Net reg outputs 9 numbers to form a $3\times3$ affine transformation matrix $A$. For T-Net reg, a regularization term $\|AA^T-I\|$ is added to the loss with weight 0.001. Batch normalization is applied to all except the last layer. Details about the network architecture can be found in the supplement. 

We trained the transformers with two state-of-the-art shape classification networks, PointNet \cite{qi2017pointnet} and Dynamic Graph CNN (DGCNN) \cite{wang2018dynamic}, and tested their performance on two datasets, Partial ModelNet40 and ScanNet Objects.

\begin{table}
    \centering
    \scriptsize
    \renewcommand\theadfont{\scriptsize}
    \begin{tabular}{cccccccc}
    \toprule
    Classifier & \multicolumn{7}{c}{PointNet} \\ \cmidrule(lr){2-8}
    Transformer & None & \multicolumn{2}{c}{T-Net} & \multicolumn{2}{c}{T-Net reg} & \multicolumn{2}{c}{IT-Net (ours)} \\
    \cmidrule(lr){2-2} \cmidrule(lr){3-4} \cmidrule(lr){5-6} \cmidrule(lr){7-8}
    \# Iterations & 0 & 1 & 2 & 1 & 2 & 1 & 2 \\
    \midrule
    Accuracy & 59.97 & 66.04 & 35.13 & 65.84 & 67.06 & 68.72 & \textbf{69.94} \\
    \midrule
    Classifier & \multicolumn{7}{c}{DGCNN} \\ \cmidrule(lr){2-8}
    Transformer & None & \multicolumn{2}{c}{T-Net} & \multicolumn{2}{c}{T-Net reg} & \multicolumn{2}{c}{IT-Net (ours)} \\
    \cmidrule(lr){2-2} \cmidrule(lr){3-4} \cmidrule(lr){5-6} \cmidrule(lr){7-8}
    \# Iterations & 0 & 1 & 2 & 1 & 2 & 1 & 2 \\
    \midrule
    Accuracy & 65.60 & 70.38 & 16.61 & 71.15 & 72.69 & 72.57 & \textbf{74.15} \\
    \bottomrule
    \end{tabular}
    \caption{Classification accuracy on Partial ModelNet40.}
    \label{tab:cls_model}
\end{table}

\begin{table}
    \centering
    \scriptsize
    \renewcommand\theadfont{\scriptsize}
    \begin{tabular}{cccccccc}
    \toprule
    Classifier & \multicolumn{7}{c}{PointNet} \\ \cmidrule(lr){2-8}
    Transformer & None & \multicolumn{2}{c}{T-Net} & \multicolumn{2}{c}{T-Net reg} & \multicolumn{2}{c}{IT-Net (ours)} \\
    \cmidrule(lr){2-2} \cmidrule(lr){3-4} \cmidrule(lr){5-6} \cmidrule(lr){7-8}
    \# Iterations & 0 & 1 & 2 & 1 & 2 & 1 & 2 \\
    \midrule
    Accuracy & 62.11 & 63.09 & 30.86 & 62.99 & 61.82 & 63.67 & \textbf{66.02} \\
    \midrule
    Classifier & \multicolumn{7}{c}{DGCNN} \\ \cmidrule(lr){2-8}
    Transformer & None & \multicolumn{2}{c}{T-Net} & \multicolumn{2}{c}{T-Net reg} & \multicolumn{2}{c}{IT-Net (ours)} \\
    \cmidrule(lr){2-2} \cmidrule(lr){3-4} \cmidrule(lr){5-6} \cmidrule(lr){7-8}
    \# Iterations & 0 & 1 & 2 & 1 & 2 & 1 & 2 \\
    \midrule
    Accuracy & 66.02 & 72.75 & 18.55 & 74.12 & 70.80 & 76.36 & \textbf{76.66} \\
    \bottomrule
    \end{tabular}
    \caption{Classification accuracy on ScanNet Objects.}
    \label{tab:cls_scan}
\end{table}

\paragraph{Results}
Table \ref{tab:cls_model} and \ref{tab:cls_scan} show the classification accuracy on Partial ModelNet40 and ScanNet Objects respectively. It can be seen that IT-Net consistently outperforms baseline transformers when trained with different classifiers. This is evidence that the advantage of IT-Net is agnostic to the classifier architecture. Further, the advantage of IT-Net over baselines on real data matches that on synthetic data. This demonstrates IT-Net's ability to process point clouds with realistic sensor noise and clutter and its potential to be incorporated in detection/pose estimation pipelines in the wild. Finally, we observe that without explicit supervision, IT-Net learns to transform the inputs into a set of canonical poses which we call ``pose clusters". The learned transformations removes pose variations among the inputs, which simplifies the classification problem. Some examples are shown in Figure \ref{fig:cls} and more visualizations are in the supplement.

We note that in the classification setting, the transformer's job is to transform the inputs into one of the pose clusters so that they can be recognized by the classifiers, which is simpler than producing precise alignments as in Sec. \ref{sec:pose}. Therefore, the performance gain diminishes as the number of unfolded iterations becomes larger than 2.

Unlike T-Net, IT-Net preserves the shape of the input without introducing any scaling or shearing. Figure \ref{fig:cls} shows that the output of T-Net is on a very different scale than the original input. This explains why the performance of T-Net drops significantly if we try to apply the same iterative scheme directly: as the network sees inputs on vastly different scales from different iterations, the training fails to converge. The regularized T-Net resolves this issue, but its performance is still worse than IT-Net.

\begin{figure*}
    \renewcommand\theadalign{bc}
    \centering
    \begin{tabular}{p{1.7cm}p{1.7cm}p{1.7cm}p{1.7cm}|p{1.7cm}p{1.7cm}p{1.7cm}p{1.7cm}}
        \thead{Input} & \thead{IT-Net\\Iteration 1} & \thead{IT-Net\\Iteration 2} & \thead{T-Net\\(scaled by 0.1)} & \thead{Input} & \thead{IT-Net\\Iteration 1} & \thead{IT-Net\\Iteration 2} & \thead{T-Net\\(scaled by 0.1)} \\
        \multicolumn{3}{c}{\includegraphics[width=5.6cm]{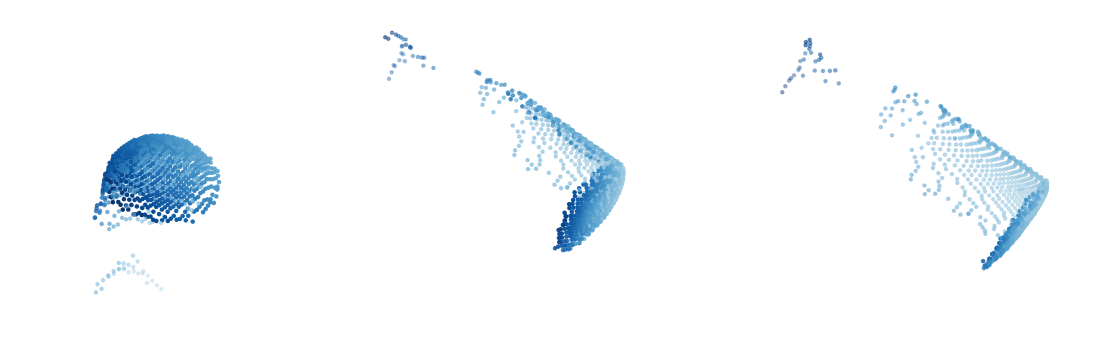}} & \includegraphics[width=1.7cm]{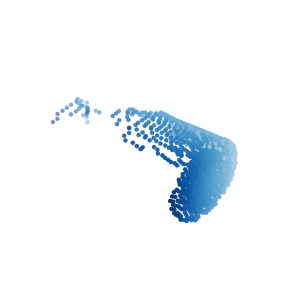} & \multicolumn{3}{c}{\includegraphics[width=5.6cm]{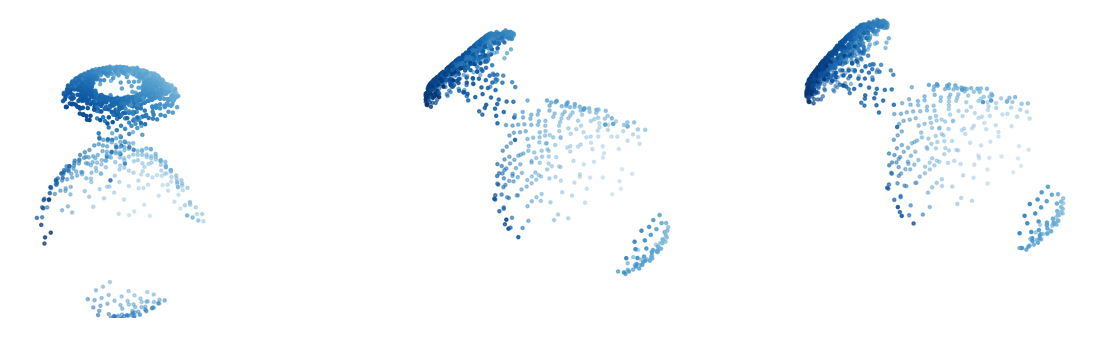}} & \includegraphics[width=1.7cm]{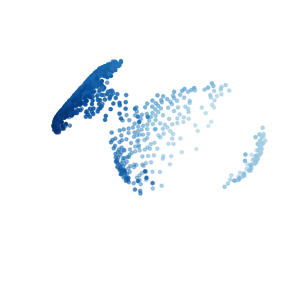} \\
        \multicolumn{3}{c}{\includegraphics[width=5.6cm]{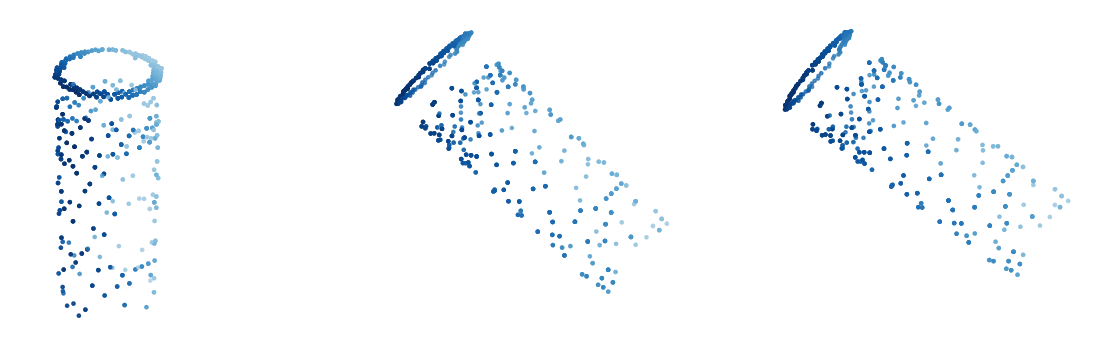}} & \includegraphics[width=1.7cm]{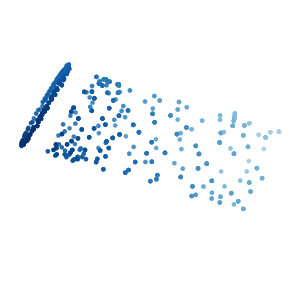} & \multicolumn{3}{c}{\includegraphics[width=5.6cm]{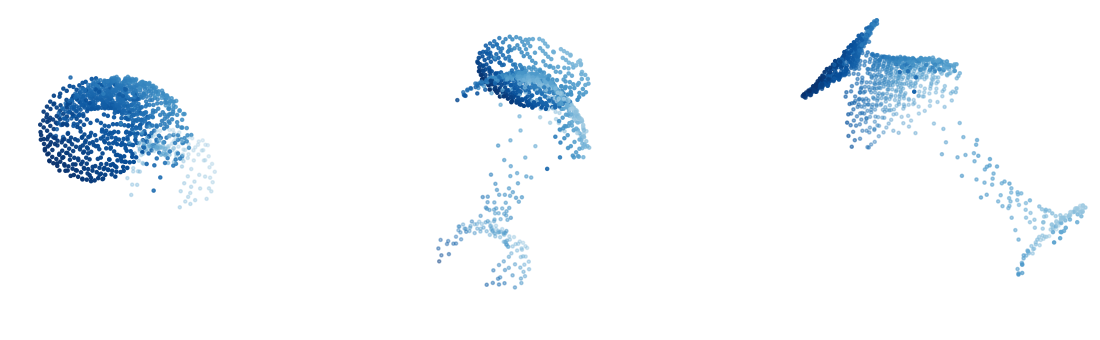}} & \includegraphics[width=1.7cm]{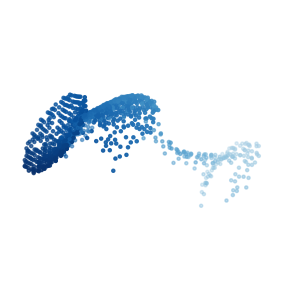} \\
    \end{tabular}
    \caption{Inputs transformed by IT-Net and T-Net trained jointly with DGCNN. Note how the input pose converges with more iterations and the similarity of final poses across different categories (columns 3, 7). T-Net's outputs are on a much different scale (10 times bigger) than the original inputs (columns 4, 8).}
    \label{fig:cls}
\end{figure*}

\begin{table*}
    \renewcommand\theadalign{tc}
    \renewcommand\theadfont{\scriptsize}
    \centering
    \scriptsize
    \begin{tabular}{c|c|cccccccccccccccc}
        \toprule
         & mean & table & chair & \thead{air\\plane} & lamp & car & guitar & laptop & knife & pistol & \thead{motor\\cycle} & mug & \thead{skate\\board} & bag & \thead{ear\\phone} & rocket & cap \\ \midrule
        \# shapes & & 5271 & 3758 & 2690 & 1547 & 898 & 787 & 451 & 392 & 283 & 202 & 184 & 152 & 76 & 68 & 66 & 55 \\ 
        \# parts & & 3 & 4 & 4 & 4 & 4 & 3 & 2 & 2 & 3 & 6 & 2 & 3 & 2 & 3 & 3 & 2 \\ \midrule
        None & 76.9 & 78.8 & 82.6 & 77.3 & 71.3 & 52.3 & 90.1 & 76.8 & 80.0 & 70.1 & 40.4 & \textbf{86.1} & 67.6 & 71.0 & 66.7 & 53.1 & 76.9 \\
        T-Net & 77.1 & 79.2 & 82.5 & 78.0 & 70.1 & 55.7 & 89.1 & 73.1 & 81.5 & 73.0 & 39.1 & 81.1 & 69.1 & \textbf{74.1} & 71.1 & 51.4 & 74.6 \\
        \midrule
        IT-Net-1 & 78.2 & 79.9 & 84.3 & 78.2 & \textbf{72.9} & 54.9 & 91.0 & \textbf{78.7} & 78.1 & 71.8 & \textbf{44.6} & 84.8 & 66.6 & 71.2 & \textbf{72.7} & \textbf{55.0} & 77.9 \\
        IT-Net-2 & \textbf{79.1} & \textbf{80.2} & \textbf{84.7} & \textbf{79.9} & 72.1 & \textbf{62.6} & \textbf{91.1} & 76.4 & \textbf{82.8} & \textbf{76.9} & 44.0 & 84.4 & \textbf{71.8} & 68.1 & 66.8 & 54.2 & \textbf{80.4} \\
        \bottomrule
    \end{tabular}
    \caption{Part segmentation results on ShapeNet Part. The number appending IT-Net indicates the number of iterations. The base segmentation model is DGCNN \cite{wang2018dynamic}. The metric is mIoU(\%) on points. The mean is calculated as the average of per-category mIoUs weighted by the number of shapes. We order the categories by number of shapes since the performance is more unstable for categories with fewer shapes.}
    \label{tab:seg}
\end{table*}

\subsection{Object Part Segmentation} \label{sec:seg}
The network used for part segmentation simply replaces the classifier in the joint transformer-classifier model from Sec. \ref{sec:cls} with a segmentation network. We use DGCNN \cite{wang2018dynamic} as the base segmentation network and compare the performance gain of adding T-Net and IT-Net. Following \cite{qi2017pointnet,wang2018dynamic}, we treat part segmentation as a per-point classification problem and train the network with a per-point cross entropy loss. Similar to Sec. \ref{sec:cls}, no explicit supervision is applied on the transformations. The networks are trained and evaluated on ShapeNet Part.

\begin{figure}[!t]
    \centering
    \begin{tabular}{p{1.6cm}p{1.6cm}p{1.6cm}p{1.6cm}}
        Original & Transformed & \centering Original & Transformed \\
        \multicolumn{2}{c}{\includegraphics[width=3.5cm]{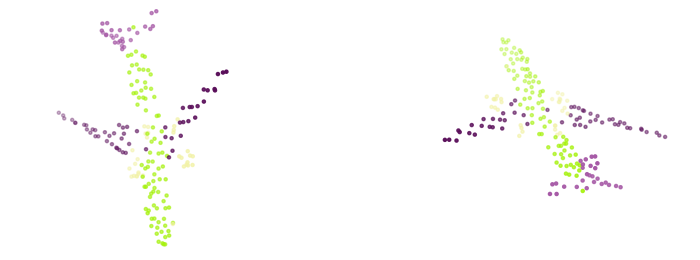}} &
        \multicolumn{2}{c}{\includegraphics[width=3.5cm]{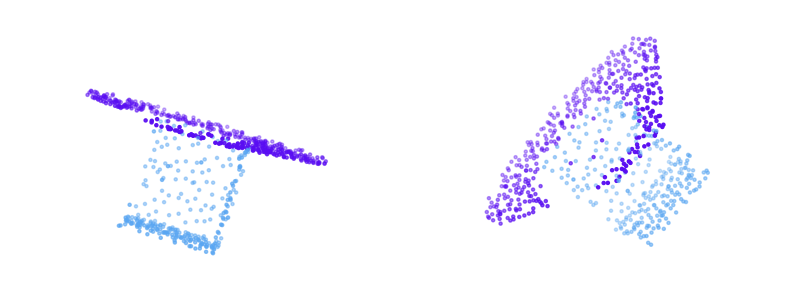}} \\
        \multicolumn{2}{c}{\includegraphics[width=3.5cm]{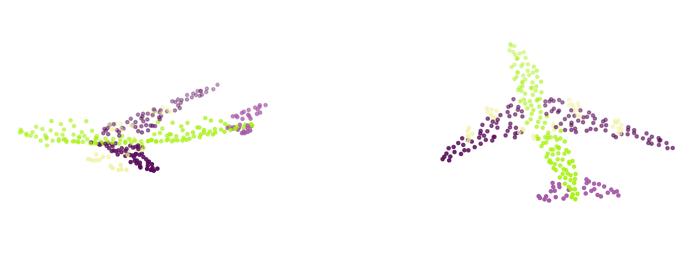}} &
        \multicolumn{2}{c}{\includegraphics[width=3.5cm]{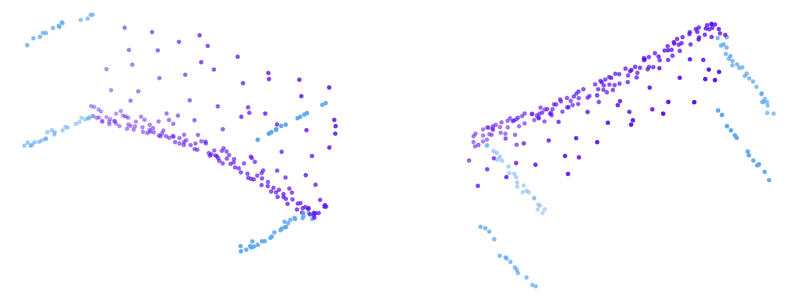}}
    \end{tabular}
    \vspace{-10pt}
    \caption{Inputs transformed by IT-Net trained with DGCNN on part segmentation. The colors indicate predictions of the segmentation network.}
    \label{fig:part_seg}
\end{figure}

\paragraph{Results}
We use the mean Intersection-over-Union (mIoU) on points as the evaluation metric following \cite{qi2017pointnet,wang2018dynamic}. The results are summarized in Table \ref{tab:seg}, which shows that IT-Net with 2 iterations outperforms other transformers in terms of mean mIoU and mIoU for most categories. Figure \ref{fig:part_seg} shows some qualitative examples. As in the case of classification, IT-Net reduces variations in the inputs caused by geometric transformations by transforming the inputs to a canonical pose. Note that the architecture of IT-Net here is identical to the ones in Sec. \ref{sec:cls}, which demonstrates the potential of IT-Net as a plug-in module for any task that requires invariance to geometric transformations without task-specific adjustments to the model architecture.

\section{Conclusion}
In this work, we propose a new transformer network on 3D point clouds named Iterative Transformer Network (IT-Net). In an iterative fashion, IT-Net outputs a rigid transformation that can be used to estimate object pose or transform the input for subsequent tasks. The effectiveness of IT-Net in various tasks shows that the classical idea of iterative refinement still applies in the context of deep learning.

IT-Net can be easily integrated with existing deep learning architectures for shape classification and segmentation, and improve the performance on these tasks with partial, unaligned inputs by introducing invariance to geometric transformations. This opens up many avenues for future research on using neural networks to extract semantic information from real world point cloud data.

\newpage
{\small
\bibliographystyle{ieee}
\bibliography{paper}
}

\newpage
\section*{Suppelementary}
\renewcommand\thesection{\Alph{section}}
\renewcommand\thesubsection{\thesection.\Alph{subsection}}
\setcounter{section}{0}

\section{Overview}
In this document, we provide technical details and visualizations in support of our paper. Here are the contents:
\begin{itemize}[noitemsep,nolistsep]
    \item[\ref{sec:data}:] details on the generation of partial point clouds;
    \item[\ref{sec:pose}:] pose estimation results on chairs in ShapeNet Pose;
    \item[\ref{sec:seg}:] part segmentation results with PointNet as base model;
    \item[\ref{sec:arch}:] details on network architecture and training;
    \item[\ref{sec:visu}:] visualizations of pose clusters learned by IT-Net.
\end{itemize}


\section{Data Generation} \label{sec:data}
In this section, we cover details on the generation of partial, unaligned object point clouds used in our experiments.

As mentioned in Sec. 4 of our paper, our dataset consists of four parts: Partial ModelNet40, ShapeNet Part, ShapeNet Pose and ScanNet Objects. The first three parts are created from scans of synthetic objects and the last part is created from real world RGB-D scans.

\begin{figure}
    \centering
    \begin{tabular}{cc|cc}
        \thead{ModelNet40} & \thead{Partial\\ModelNet40} & \thead{ModelNet40} & \thead{Partial\\ModelNet40} \\
        \multicolumn{2}{c|}{\includegraphics[width=0.42\linewidth]{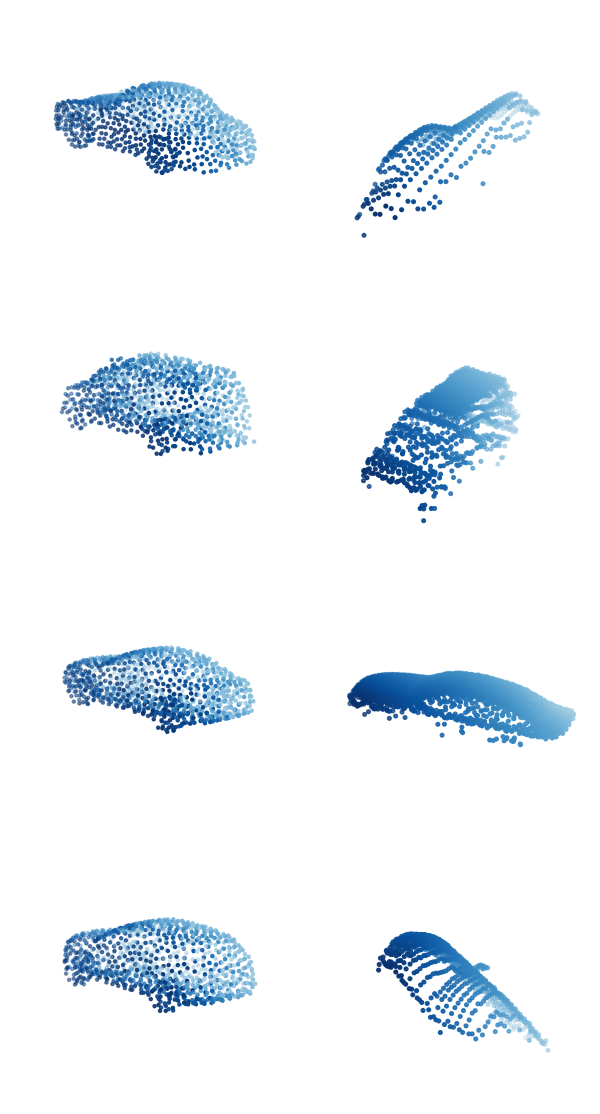}} & \multicolumn{2}{c}{\includegraphics[width=0.42\linewidth]{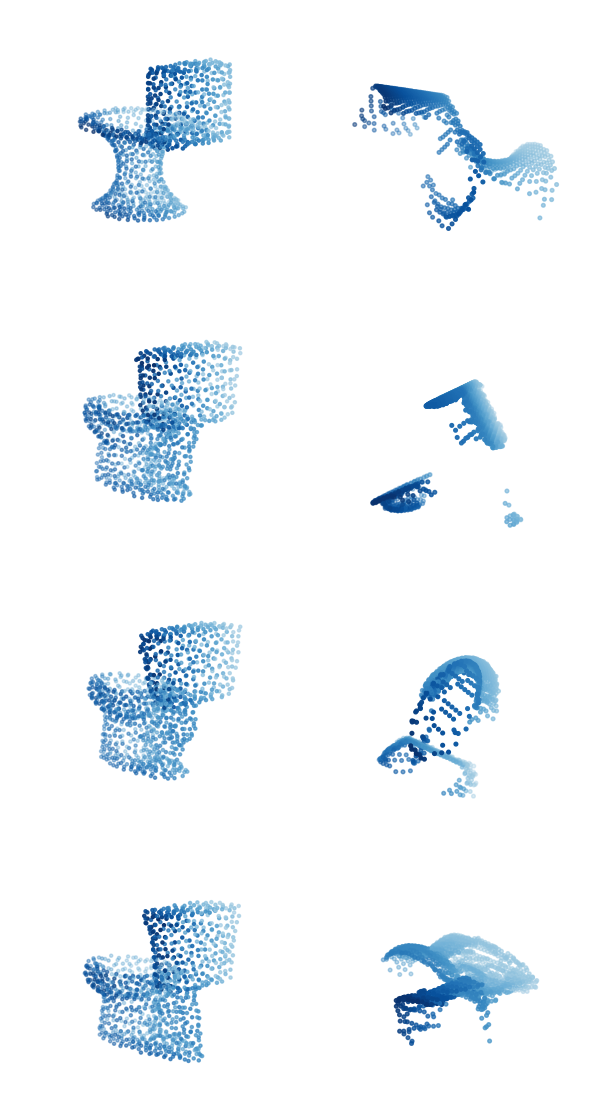}}
    \end{tabular}
    \caption{Comparison between ModelNet40 used in \cite{qi2017pointnet,wang2018dynamic} and partial ModelNet40 used in our experiments.}
    \label{fig:data}
\end{figure}

In Partial ModelNet40, each point cloud is generated by fusing a sequence of depth scans of CAD models from the aligned ModelNet40 dataset \cite{sedaghat2016orientation}. The fused point clouds are in the coordinates of the first camera. The orientation of the first camera is uniformly random and the distance between the camera center and the object center is uniform between 2 to 4 units (the models are normalized into the unit sphere). Subsequent camera positions are generated by rotating the camera around the object center for up to 30 degrees. We use Blender \cite{blender} to render the depth scans. Compared to the uniformly sampled point clouds used to evaluate classification in prior works \cite{qi2017pointnet,wang2018dynamic}, our dataset contains much more challenging inputs with various poses, densities and levels of incompleteness (see Figure \ref{fig:data}).

For ShapeNet Part, since the part labels \cite{yi2016scalable} are associated with sampled point clouds instead of the original mesh models from ShapeNet \cite{chang2015shapenet}, we generate partial point clouds by virtually scanning the point clouds. Specifically, we randomly rotate the complete point cloud and project the points onto a virtual image with pixel size 0.02 in the $xy$-plane. For each pixel of the virtual image, we keep the point with the smallest $z$ value and discard all other points that project onto the same pixel as they are considered as occluded by the selected point. This procedure mimics an orthographic depth camera and creates partial point clouds that look much like those created from rendered depth scans.

\begin{table}
    \centering
    \footnotesize
    \renewcommand\theadfont{\footnotesize}
    \begin{tabular}{c|ccc}
        \toprule 
        & \thead{Partial\\ModelNet} & \thead{ShapeNet\\Part} & \thead{ShapeNet\\Pose} \\
        \midrule
        Task & Classification & Segmentation & Pose estimation \\
        Source & ModelNet40 \cite{sedaghat2016orientation} & ShapeNet \cite{chang2015shapenet} & ShapeNet \cite{chang2015shapenet} \\
        \# classes & 40 & 16 & 1 \\
        \# train & 78,744 & 12,137 & 22,000 \\
        \# test & 2,468 & 1,870 & 2,000 \\
        \# scans & 1-4 & 1 & 1 \\
        Scan size & $64\times64$ & $50\times50$ & $128\times128$ \\
        Focal length & 57 & $\infty$ & 64 \\
        \thead{Distance to\\object center} & 2-4 & $\infty$ & 1-2 \\
        \bottomrule
    \end{tabular}
    \caption{Statistics and parameters of our partial point cloud dataset for classification, segmentation and pose estimation.}
    \label{tab:data}
\end{table}

\begin{figure*}
    \centering
    \includegraphics[width=0.8\linewidth]{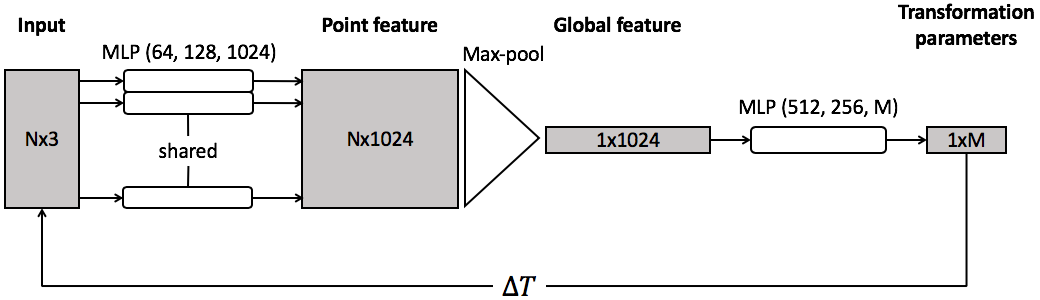}
    \caption{Detailed transformer architecture. Numbers in the parenthesis indicate the number of neurons in each MLP layer. The output dimension $M$ is 7 for IT-Net and 9 for T-Net and T-Net reg.}
    \label{fig:arch}
\end{figure*}

\begin{table*}
    \renewcommand\theadalign{tc}
    \renewcommand\theadfont{\scriptsize}
    \centering
    \scriptsize
    \begin{tabular}{c|c|cccccccccccccccc}
        \toprule
         & mean & table & chair & \thead{air\\plane} & lamp & car & guitar & laptop & knife & pistol & \thead{motor\\cycle} & mug & \thead{skate\\board} & bag & \thead{ear\\phone} & rocket & cap \\ \midrule
        \# shapes & & 5271 & 3758 & 2690 & 1547 & 898 & 787 & 451 & 392 & 283 & 202 & 184 & 152 & 76 & 68 & 66 & 55 \\ 
        \# parts & & 3 & 4 & 4 & 4 & 4 & 3 & 2 & 2 & 3 & 6 & 2 & 3 & 2 & 3 & 3 & 2 \\ \midrule
        None & 67.9 & 71.6 & 75.2 & 68.8 & 56.9 & 48.2 & 82.4 & 58.0 & 68.5 & 61.7 & 39.0 & 65.6 & 49.6 & 41.9 & 43.5 & 28.1 & \textbf{50.9} \\
        T-Net & 71.1 & 73.7 & 77.5 & 73.6 & 60.2 & 53.0 & 85.8 & \textbf{63.2} & 73.6 & 65.4 & 48.5 & 70.3 & 57.7 & 15.9 & 41.8 & 41.7 & 48.5 \\
        \midrule
        IT-Net-1 & 72.3 & 74.5 & \textbf{78.7} & 75.9 & 60.6 & \textbf{57.7} & 85.1 & 58.3 & \textbf{78.6} & 67.9 & \textbf{51.5} & 70.3 & 61.6 & 31.6 & \textbf{53.9} & 35.2 & 45.3 \\
        IT-Net-2 & \textbf{72.6} & \textbf{75.1} & 78.3 & \textbf{76.3} & \textbf{62.1} & 56.3 & \textbf{86.8} & 58.9 & 74.5 & \textbf{68.6} & 46.4 & \textbf{70.6} & \textbf{65.9} & \textbf{43.5} & 51.6 & \textbf{42.6} & 45.9 \\
        \bottomrule
    \end{tabular}
    \caption{Part segmentation results on partial shapes from ShapeNet Part. The number appending IT-Net indicates the number of unfolded iterations during training. The base segmentation model is PointNet \cite{qi2017pointnet}. The metric is mIoU(\%) on points. The mean is calculated as the average of per-category mIoUs weighted by the number of shapes.}
    \label{tab:seg}
\end{table*}

The partial point clouds in ShapeNet Pose are created from ShapeNet models in a similar way as Partial ModelNet40, except that the label is the transformation between the camera coordinates and the model coordinates instead of the category. Table \ref{tab:data} summarizes the statistics and parameters used to generate the synthetic parts of our dataset.

The point clouds in ScanNet Objects are created from 1,512 real world RGB-D scans of indoor scenes in ScanNet \cite{dai2017scannet}. In total, we collect 9,122 object points clouds by cropping the scans with labeled bounding boxes, where sensor noise and clutter in the boxes are kept in the resulting point clouds. We normalize the point clouds into the unit sphere and translate their centroids to the origin.

\begin{table}
    \centering
    \small
    \begin{tabular}{cccccccc}
        \toprule
        \thead{Unfolded \\ iterations} & 1 & 2 & 3 & 4 & 5 & 6 & 7 \\
        \midrule
        \thead{Pose \\ accuracy} & 27.4 & 36.3 & 43.6 & 62.4 & 64.3 & 56.4 & 51.1 \\
        \bottomrule
    \end{tabular}
    \caption{Pose accuracy (\%) with error threshold $10^\circ,0.1$ of IT-Nets on chair point clouds in ShapeNet Pose.}
    \label{tab:pose}
\end{table}

\section{Pose Estimation on Chairs} \label{sec:pose}
As mentioned in Sec. 4.1 of our paper, we performed experiments on the chair point clouds in ShapeNet Pose using the same setting as on the car point clouds. The results verified our claims in Sec. 4.1. First, Table \ref{tab:pose} shows that IT-Net trained with 5 unfolded iterations gives the best performance, which indicates that the curriculum generated by 5-iteration IT-Net strikes a balance between examples with small and large pose errors. Second, Figure \ref{fig:pose} shows that on a different category, IT-Net keeps the property that pose accuracy increases with the number of testing iterations.
\begin{figure}
    \centering
    \includegraphics[width=0.85\linewidth]{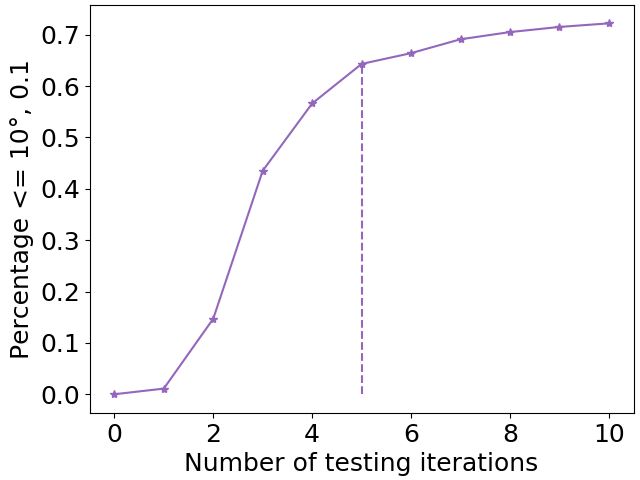}
    \caption{Pose accuracy (\%) against the number of iterations applied during inference. The dotted line corresponds to the number of unfolded iterations in training.}
    \label{fig:pose}
\end{figure}

\section{Part Segmentation with PointNet} \label{sec:seg}
Table \ref{tab:seg} shows part segmentation results on ShapeNet Part using PointNet \cite{qi2017pointnet} as the base segmentation model instead of DGCNN \cite{wang2018dynamic} used in the experiments in Sec. 4.3. Similar to the classification results in Sec. 4.2, the segmentation results show evidence that the advantage of IT-Net is agnostic to the architecture of the segmentation network.

\section{Architecture and Training Details} \label{sec:arch}
Figure \ref{fig:arch} shows the detailed architecture of the PointNet \cite{qi2017pointnet} used as the pose regression network in all our experiments. The architecture consists of three parts. The first part is a multi-layer perceptron (MLP) that is applied on each point independently. It takes the $N\times3$ coordinate matrix and produces a $N\times1024$ feature matrix. The second part is a max-pooling function which aggregates the features in to a $1\times1024$ vector. The third part is another MLP that regresses $M$ pose parameters from the 1024-dimensional global feature vector. We have $M=7$ for IT-Net and $M=9$ for T-Net and T-Net-reg.

We use publicly available implementations of PointNet \cite{qi2017pointnet} and DGCNN \cite{wang2018dynamic} for the classification and segmentation networks. The detailed network architectures can be found in Section C of the supplementary for \cite{qi2017pointnet} and Section 5.1 and 5.4 of \cite{wang2018dynamic}.

The pose estimation networks in Sec. 4.1 are trained for 20000 steps with batch size 100. We use Adam optimizer with an initial learning rate of 0.001, decayed by 0.7 every 2000 steps. The initial decay rate for batch normalization is 0.5 and gradually increased to 0.99.

The joint transformer-classification networks in Sec. 4.2 are trained for 50 epochs with batch size 32. We use the Adam optimizer with an initial learning rate of 0.001, decayed by 0.7 every 6250 steps. The initial decay rate for batch normalization is 0.5 and gradually increased to 0.99. We clip the gradient norm to 30.

The joint transformer-segmentation networks in Sec. 4.3 are trained for 200 epochs with batch size 32. Other hyperparameters are the same as in Sec. 4.2.

\section{Pose Cluster Visualizations} \label{sec:visu}
\begin{figure*}
    \centering
    \begin{subfigure}{0.87\linewidth}
        \begin{tabular}{p{0.3\linewidth}p{0.3\linewidth}p{0.3\linewidth}}
            \centering Input (Iteration 0) & \centering Iteration 1 & \centering Iteration 2
        \end{tabular}
        \includegraphics[width=\linewidth]{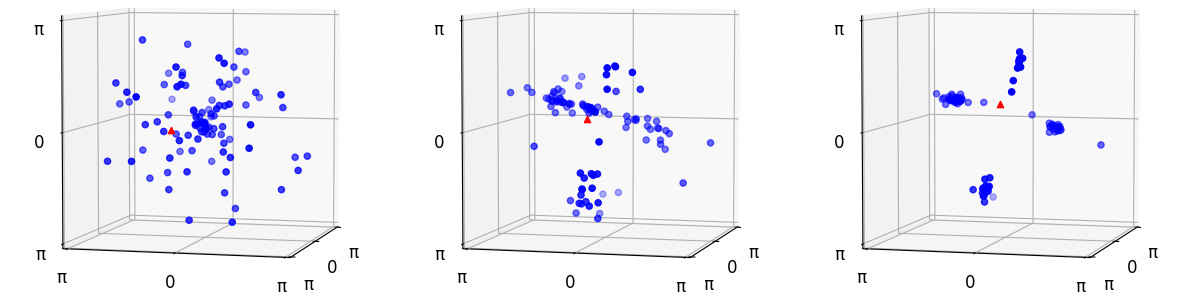}
        \caption{Distribution of orientations at different iterations.}
        \label{fig:guitar_clus}
    \end{subfigure}
    \begin{subfigure}{0.12\linewidth}
        \includegraphics[width=\linewidth]{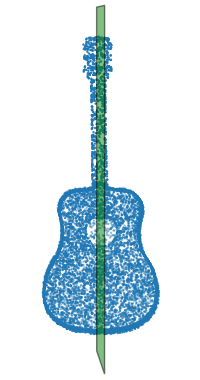}
        \caption{Reflection symmetry of guitars.}
        \label{fig:guitar_sym}
    \end{subfigure}
    \begin{subfigure}{0.48\linewidth}
        \raisebox{0.1cm}{\includegraphics[width=\linewidth]{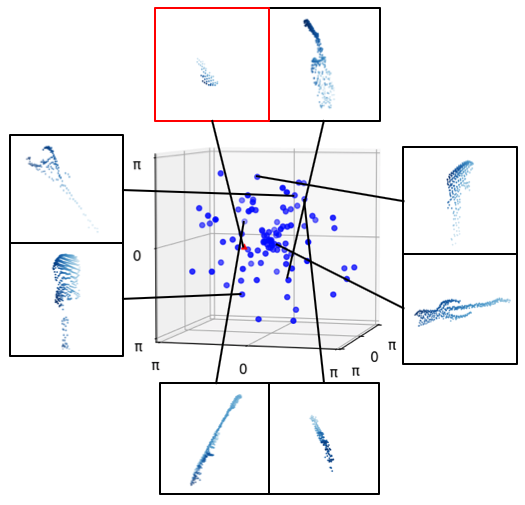}}
        \caption{Examples of original inputs (Iteration 0).}
        \label{fig:guitar_orig}
    \end{subfigure}
    \begin{subfigure}{0.51\linewidth}
        \includegraphics[width=\linewidth]{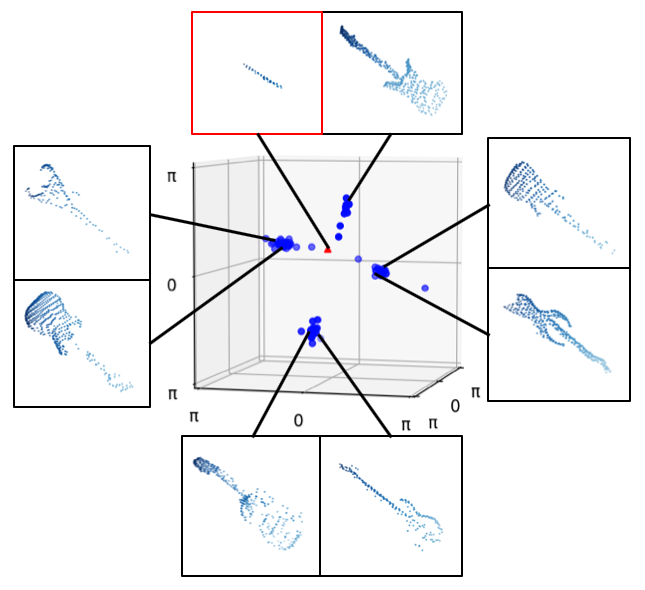}
        \caption{Examples of transformed inputs (Iteration 2).}
        \label{fig:guitar_trans}
    \end{subfigure}
    \caption{Pose cluster visualization for guitars. \textbf{(a)} Distribution of axis-angle representation of orientations of all test examples at different iterations. Note how clusters emerge from uniformly distributed poses. Correctly classified examples are shown in blue and incorrectly classified examples are shown in red. \textbf{(b)} The reflection symmetry present in most guitars. \textbf{(c)} Examples of original inputs at iteration 0. The object orientations are uniformly distributed. \textbf{(d)} Examples of transformed inputs at iteration 2. Note that these are the inputs received by the classifier. The object orientations are grouped into 4 clusters, but visually there seems to be only 2 major orientations due to the reflection symmetry shown in \textbf{(b)}. A failure case caused by heavy occlusion is shown in the red box.}
    \label{fig:guitar}
\end{figure*}

\begin{figure*}
    \centering
    \begin{subfigure}{0.87\linewidth}
        \begin{tabular}{p{0.3\linewidth}p{0.3\linewidth}p{0.3\linewidth}}
            \centering Input (Iteration 0) & \centering Iteration 1 & \centering Iteration 2
        \end{tabular}
        \includegraphics[width=\linewidth]{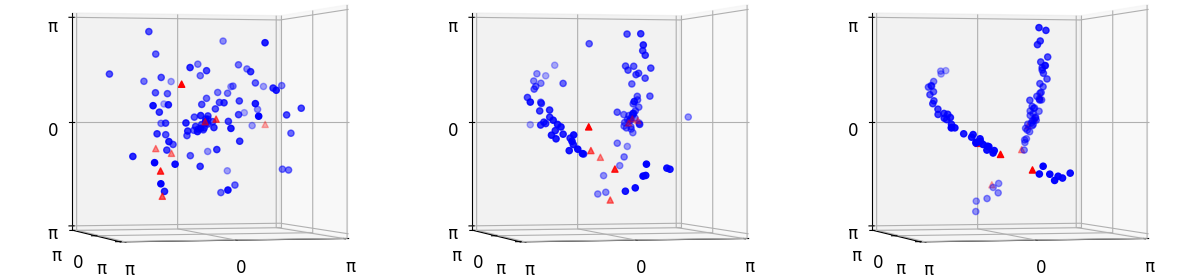}
        \caption{Distribution of orientations at different iterations.}
        \label{fig:bottle_clus}
    \end{subfigure}
    \begin{subfigure}{0.12\linewidth}
        \raisebox{0.2cm}{\includegraphics[width=\linewidth]{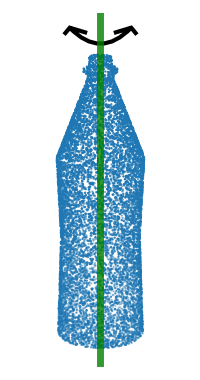}}
        \caption{Rotational symmetry of bottles.}
        \label{fig:bottle_sym}
    \end{subfigure}
    \begin{subfigure}{0.48\linewidth}
        \includegraphics[width=\linewidth]{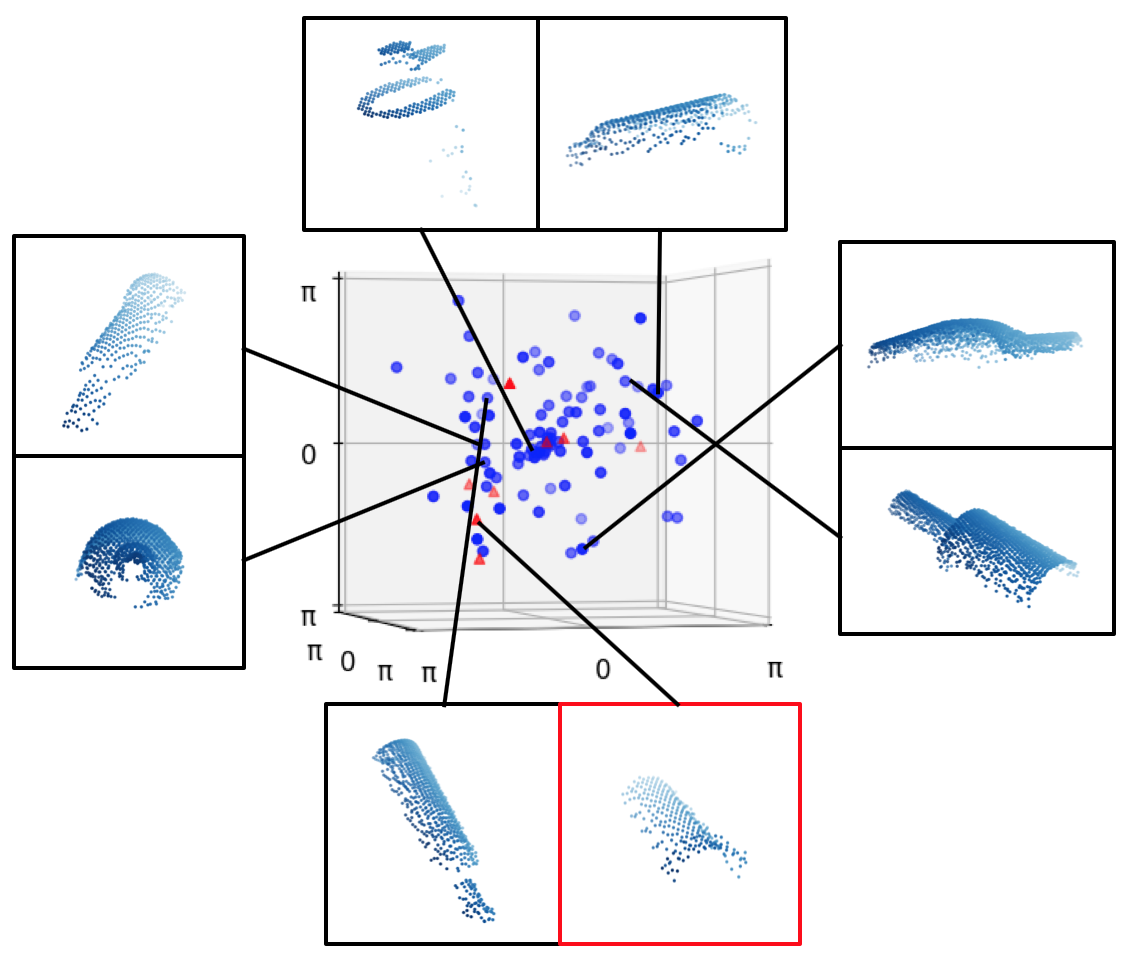}
        \caption{Examples of original inputs (Iteration 0).}
        \label{fig:bottle_orig}
    \end{subfigure}
    \begin{subfigure}{0.5\linewidth}
        \includegraphics[width=\linewidth]{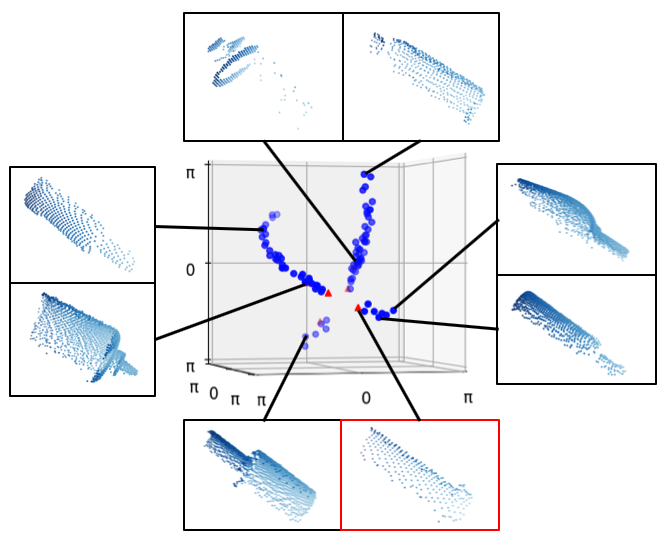}
        \caption{Examples of transformed inputs (Iteration 2).}
        \label{fig:bottle_trans}
    \end{subfigure}
    \caption{Pose cluster visualization for bottles. \textbf{(a)} Distribution of axis-angle representation of orientations of all test examples at different iterations. Note how clusters emerge from uniformly distributed poses. Correctly classified examples are shown in blue and incorrectly classified examples are shown in red. \textbf{(b)} The rotational symmetry present in most bottles. \textbf{(c)} Examples of original inputs at iteration 0. The object orientations are uniformly distributed. \textbf{(d)} Examples of transformed inputs at iteration 2. Note that these are the inputs received by the classifier. The object orientations after transformation are grouped into 2 clusters. The clusters have semicircle shapes since any orientation in these semicircles are in fact indistinguishable due to the rotational symmetry shown in \textbf{(b)}. A failure case is shown in the red box. In this case the model misclassifies the bottle as a vase.}
    \label{fig:bottle}
\end{figure*}

Figure \ref{fig:guitar} and Figure \ref{fig:bottle} show visualizations of the ``pose clusters" learned by IT-Net as mentioned in Sec. 4.2. To visualize the pose clusters, we calculate the difference between the canonical orientation of the input shape and the orientation of the transformed input at different iterations. Then, we convert the orientation difference into axis-angle representation, which is a 3D vector, and plot these vectors for all test examples in a particular category. The learned pose clusters for the guitar and the bottle category are shown in Figure \ref{fig:guitar} and Figure \ref{fig:bottle} respectively. The network being visualized is a 2-iteration IT-Net trained with DGCNN for shape classification on Partial ModelNet40.

We observe that although the object poses are uniformly distributed initially, clusters of poses emerge after applying the transformations predicted by IT-Net (Figure \ref{fig:guitar_clus}, \ref{fig:bottle_clus}). This is evidence that IT-Net discover a canonical space to align the inputs with no explicit supervision. Interestingly, there are usually more than one cluster and the shapes of the clusters are related to the symmetries of the object (Figure \ref{fig:guitar_sym}, \ref{fig:bottle_sym}). Further, we note that sometimes even objects across different categories are aligned after being transformed by IT-Net (Figure \ref{fig:guitar_trans}, \ref{fig:bottle_trans}). 

\setlength{\belowcaptionskip}{0pt}

\end{document}